%% file: main.tex
\documentclass{vgtc}                          

\ifpdf
  \pdfoutput=1\relax                   
  \usepackage{graphicx}                
  \DeclareGraphicsExtensions{.pdf,.png,.jpg,.jpeg} 
\else
  \usepackage{graphicx}                
  \DeclareGraphicsExtensions{.eps}     
\fi%

\graphicspath{{figures/}{pictures/}{images/}{./}} 

\usepackage{microtype}                 
\PassOptionsToPackage{warn}{textcomp}  
\usepackage{textcomp}                  
\usepackage{mathptmx}                  
\usepackage{times}                     
\usepackage{cite}                      
\usepackage{tabu}                      
\usepackage{booktabs}                  

\usepackage{caption}
\usepackage{makecell}
\usepackage{subcaption}
\usepackage{amsfonts}
\usepackage{enumitem}
\usepackage{xcolor}
\usepackage{amsmath}
\usepackage{url}

\newcommand{\myparagraph}[1]{\noindent\textbf{#1.}}

\onlineid{1127}

\vgtccategory{Research}

\title{Model-aware 3D Eye Gaze from Weak and Few-shot Supervisions}

\author{
Nikola Popovic$^{1}$$\thanks{Equal contribution. Also, corresponding authors with the following e-mails: nipopovic@vision.ee.ethz.ch, dchristodoul@student.ethz.ch}$ \and 
Dimitrios Christodoulou$^{1}$\footnotemark[1] \and  
Danda Pani Paudel$^{1,2}$ \and  
Xi Wang$^{1}$ \and
Luc Van Gool$^{1,2}$ \and
$^{1}$Computer Vision Laboratory, ETH Zurich, Switzerland\\
$^{2}$INSAIT, Sofia University, Bulgaria\\
}

\teaser{
  \centering
  \includegraphics[width=1\linewidth]{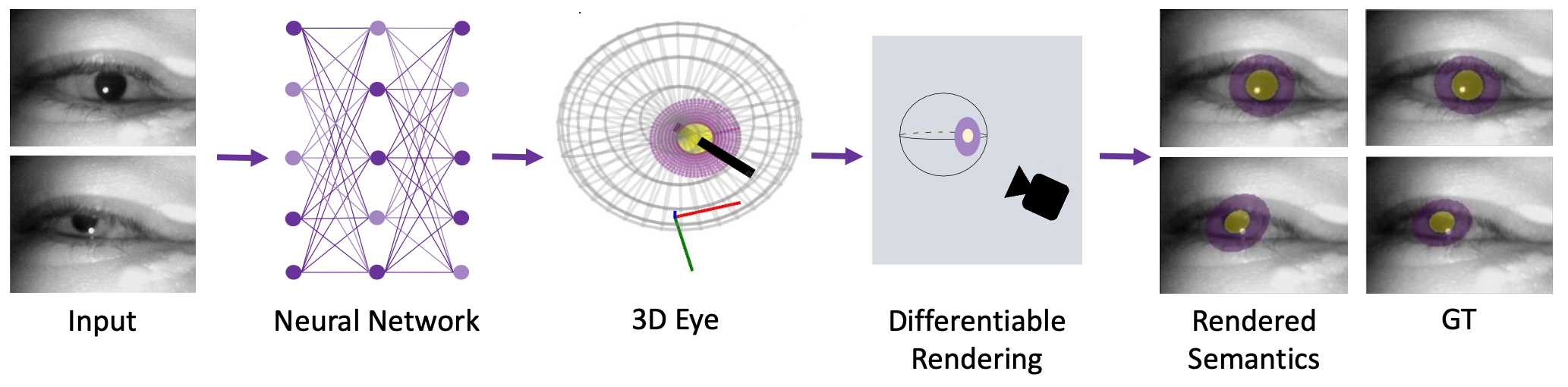}
  \caption{We propose a hybrid approach that outputs 3D eye model, eye semantic segmentation, camera intrinsic and pose. We use only 2D eye semantic segmentation masks and fewer 3D gaze labels for supervision. Left: the proposed learning framework; right: obtained segmentation examples compared to ground truth (GT) semantic masks.  }
  \label{fig:teaser}
}

\input{latex/0_abstract}

\CCScatlist{
  \CCScatTwelve{Gaze tracking}{Eye segmentation}{Eye model}
}

\begin{document}

\input{latex/1_introduction}

\input{latex/2_related_works}
\input{latex/3_method}

\input{latex/4_experiments}

\input{latex/5_conclusion}

\acknowledgments{This research was co-financed by Innosuisse under the project Know Where To Look, Grant No. 59189.1 IP-ICT. The aforementioned project was a research collaboration between the Computer Vision Lab at ETH Zurich and Aegis Rider AG.}

\bibliographystyle{abbrv-doi}

\bibliography{main}
\end{document}

%% file: latex/0_abstract.tex
\abstract{
The task of predicting 3D eye gaze from eye images can be performed either by (a) end-to-end learning for image-to-gaze mapping or by (b) fitting a 3D eye model onto images. The former case requires 3D gaze labels, while the latter requires eye semantics or landmarks to facilitate the model fitting. Although obtaining eye semantics and landmarks is relatively easy, fitting an accurate 3D eye model on them remains to be very challenging due to its ill-posed nature in general. On the other hand, obtaining large-scale 3D gaze data is cumbersome due to the required hardware setups and computational demands.

In this work, we propose to predict 3D eye gaze from weak supervision of eye semantic segmentation masks and direct supervision of a few 3D gaze vectors. The proposed method combines the best of both worlds by leveraging large amounts of weak annotations--which are easy to obtain, and only a few 3D gaze vectors--which alleviate the difficulty of fitting 3D eye models on the semantic segmentation of eye images. Thus, the eye gaze vectors, used in the model fitting, are directly supervised using the few-shot gaze labels. Additionally, we propose a transformer-based network architecture, that serves as a solid baseline for our improvements. Our experiments in diverse settings illustrate the significant benefits of the proposed method, achieving about $5 ^{\circ}$ lower angular gaze error over the baseline, when only $0.05\%$ 3D annotations of the training images are used.
The source code is available at ~\url{https://github.com/dimitris-christodoulou57/Model-aware_3D_Eye_Gaze}.
} 

%% file: latex/1_introduction.tex

\firstsection{Introduction}

\maketitle
Knowledge about the human eye gaze is instrumental in behavior analysis, affective computing, human-AI interactions, and eXtended Reality~\cite{hansen2009eye,jacob2003eye,schwartz2020eyes, plopski2022eye}. Often in many applications such as eye-gaze assistive technology~\cite{corno2002cost}, it is necessary to know where exactly is being looked at. 
 Such reasoning requires a precise 3D eye gaze vector in world coordinates. The 3D gaze in the coordinate frame of the eye-looking camera can be mapped to the world, using the known pose of the very same camera. In this context, we are interested to estimate the 3D gaze with respect to the eye-camera coordinates. 

One intuitive way of extracting 3D gaze from eye images is via model fitting~\cite{swirski2013fully}, where a 3D parametric eyeball model is reconstructed from image observations.   
Once the eye model is fitted on the images, it is straightforward to obtain the sought 3D gaze. The difficulty of accurately fitting the model is primarily due to variations in viewpoints which often lead to occlusions, 
ill-posed nature of reconstructing 3D models from images where only eye semantics are available, 
and changes in lighting conditions.  
Another alternative is to use end-to-end learning, where the mapping from image to gaze is learned using paired examples. We wish to augment the learning-based gaze prediction methods by making use of geometric constraints imposed by modeling the 3D eyes. 
Such model-award gaze prediction methods can take advantage of both model-based and learning-based methods, and in turn, may allow us to learn gaze prediction using very few gaze labels. 

In this work, we make a practical consideration that obtaining the semantic labels of eye parts is relatively easy compared to obtaining detailed geometric models of the eyeballs. Thanks to the ease of manual annotation many such public datasets exist~\cite{fuhl2021teyed, garbin2019openeds}. However, these semantics only serve as a weak supervisory signal for learning 3D eye gaze estimation. On the other hand, collecting 3D gaze labels required for direct supervision is cumbersome, mainly due to the requirements of the hardware and computational setups~\cite{fuhl2021teyed,yan2022dataset,hanhart2014eyec3d}. For example, a common way of obtaining 3D gaze labels is by asking the user to look at a known 3D target, where the location of the 3D target point and the pose of the calibrated eye camera needs to be known in a common coordinate system. This often requires tracking the human head and calibrating the eye camera with respect to it. In some device-specific settings, the head-to-eye-camera calibration may be avoided at the cost of making the estimations device-dependent. Besides the device-specific calibrations, user-specific information (e.g. distance between eyeball center to camera) may also be required~\cite{wang20183d}. These complications make the collection of 3D gaze data very tedious. Therefore, we wish to make use of only a few images whose corresponding 3D gaze directions are known, while relying on the weak supervision from semantics. The few-shot examples of 3D gaze vectors are expected to alleviate the ambiguity of model fitting on semantics only. 

A major technical challenge addressed in this paper is on making the eye model differentiable so that it can be used in an end-to-end learning framework. In particular, we propose a method to estimate the 3D eye model parameters using available eye semantics in a weakly-supervised manner. This is carried out by performing a sequence of differential operations on 3D points sampled from the canonical eye model whose semantic labels are known (from both iris and pupil). These operations provide us the mapping of 3D points 
onto the 2D image, as a function of eye model and camera parameters. The projected 3D points 
can now be compared against the image segmentation masks for weak supervision, and  
3D gaze directions can be directly derived from the reconstructed 3D eyes. 
A high-level overview of the proposed method is depicted in Figure~\ref{fig:teaser}.

\begin{figure*}[th!]
    \centering
    \includegraphics[width=0.95\textwidth]{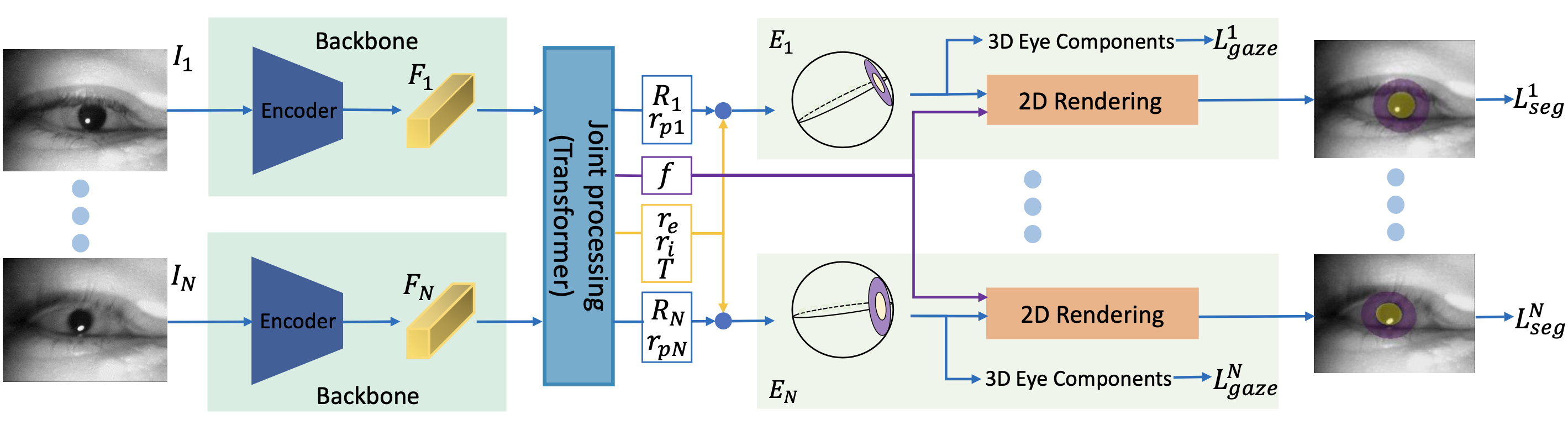}
    \vspace{-5pt}
    \caption{We process a sequence of images to predict frame-dependent and independent parameters. Each frame is first embedded using the backbone feature extractor and then jointly processed using a transformer network to produce 3D eye and camera parameters. The 3D gaze labels directly supervise predictions, while semantic masks supervise the rendered semantic regions obtained using the proposed method. }
    \vspace{-5pt}
    \label{fig: Flow}
\end{figure*}

We process a sequence of video frames to predict per-frame varying rotation and pupil's radius, respectively, to consider the rotating eyeball and dilating/contracting pupil. The remaining four variables, radii of iris and eyeball, camera's translation, and focal length, are considered to be constant throughout the same video sequence. To facilitate the learning from multiple frames, we perform joint processing of the visual features obtained from the individual frames, using a transformer-based architecture. 
Our experiment shows that the proposed architecture is powerful when supervised fully by a large set of 3D gaze labels. 
The proposed method significantly outperforms the compared methods in a few-shot setting, highlighting the benefits of such a hybrid approach. 
To this end, we argue that the ability to provide accurate 3D gaze estimations in the few-shot setting has an important implication in practice. For example, it would enable an easy adaptation to individual differences through a simple personalized calibration procedure.    
The proposed method can ease the adaptation process by needing significantly fewer gaze labels.
The major contributions of this paper can be summarized as:
\begin{itemize}[noitemsep, topsep=0pt]
    \item We propose a differentiable eye model which allows us to jointly learn from semantics and gaze labels. 
    \item We propose a transformer-based architecture that jointly processes multiple consecutive frames, allowing us to estimate per-frame varying parameters while keeping the underlying 3D eyeball model consistent.  
    \item Our method achieves significant improvements over the state-of-the-art method on the TEyeD dataset, reducing the estimated 3D gaze error from $7.1^{\circ}$ to $0.96^{\circ}$. 
    \item In the few-shot setting, the proposed method achieves about $5^{\circ}$ lower angular gaze error over the baseline, when only $0.05\%$ 3D annotations are used for training.
\end{itemize}

%% file: latex/2_related_works.tex
\section{Related Works}

\myparagraph{Eye Segmentation} Eye segmentation refers to the task of identifying different eye regions of an image that pertains specifically to the eyes. It usually includes identifying the parts of the pupil (the dark center), iris (the color area surrounding the pupil), and sclera (the white region of the eyes). 
Eye segmentation is important for various applications such as gaze analysis, biometrics and authentification, medical diagnosis, and AR/VR/XR.
Early studies rely on classical image processing methods such as edge detection~\cite{haro2000detecting}, contour extraction~\cite{martinikorena2018fast}, and ellipse fitting~\cite{swirski2012robust}.  
Recent work relies on deep neural networks to segment different eye regions by utilizing U-Net~\cite{ronneberger2015u, lozej2018end}, Feature Pyramid Network~\cite{lin2017feature}, Mask R-CNN~\cite{he2017mask} and encoder-decoder network~\cite{rot2018deep}.

\myparagraph{Appearance-Based Gaze Estimation} Enabled by recently released large-scale gaze prediction datasets~\cite{krafka2016eye, zhang2015appearance, zhang2020eth}, modern appearance-based gaze estimation methods use deep learning models to predict gaze directions from images. 
Either cropped eye images~\cite{zhang2015appearance}, face images~\cite{zhang2017s, balim2023efe}, or a combination of both~\cite{krafka2016eye, park2020towards} can be used as input, and deep features extracted from the input are then used to predict gaze directions. The recent GazeOnce method~\cite{zhang2022gazeonce} relies on multi-task learning to output facial landmarks, face location along with gaze directions. A self-supervised approach based on contrastive learning has been proposed for domain adaptation~\cite{wang2022contrastive}. Through data augmentation, features with close gaze labels are pulled together while features with different gaze labels are pushed apart. Multi-view data has also been used to learn consistent feature representations in an unsupervised manner~\cite{gideon2022unsupervised}, forcing appearance consistency across different views. 

\myparagraph{Model-Based Gaze Estimation} We refer to the gaze estimation methods which reconstruct a 3D parametric eyeball model as model-based methods. Such approaches can in general capture subject-specific eyeball features such as the eyeball radius, the corner region, and the pupil size.   
Conventional approaches~\cite{chen2008robust, hennessey2006single} rely on the detection and tracking of glints in infrared images, which are the reflection of light sources on the cornea. 
A simplified glint-free 3D eyeball model was proposed in~\cite{swirski2013fully} where images from a single camera were taken as input. The algorithm fits the pupil motion observed in the images and the obtained eyeball model can then be directly used to calculate the gaze vectors. 
Corneal refraction is considered in follow-up works~\cite{dierkes2018novel, dierkes2019fast} to further improve the estimation accuracy. 
RGB-D images have been used to fit more subject-specific eyeball models by introducing more parameters~\cite{wang2016real} and a recent study uses stereo images to construct deformable eyeball models~\cite{kuangtowards}.   

Model-based methods in general have better generalization ability, however, they are not differentiable and thus cannot transfer knowledge to new device setups (e.g. a glint-based method needs to be completely redesigned for a glint-free setup). 
In contrast, appearance-based learning approaches can extract better eye features, however, they require sufficient gaze labels which are difficult to obtain for a new setup. 
We combines the advantage of both model-based and appearance-based gaze estimation approaches and predict gaze directions by predicting a fully differentiable 3D eyeball model which can additionally be weakly supervised with semantics.



%% file: latex/3_method.tex
\section{Method}
Our method jointly processes a sequence of consecutive eye video frames $\mathcal{I} = \{I_1, ..., I_n, ..., I_N\}$ and estimate their 3D eye model parameters, which allow us to obtain the corresponding 3D gaze vectors and semantics. The first step in our pipeline is to independently process each eye frame $I_n$ and extract its global features. Afterward, all features from the sequence are jointly processed to estimate the 3D eye parameters, which are then are used to deform the canonical eye model and transform it in the camera coordinates. 
The 3D eye model can be rendered onto the image plane, providing the 2D semantic masks of the pupil and iris. 
Note that all the above-mentioned steps are fully differentiable, and the rendered 2D semantic masks can be supervised with semantic labels, thus providing supervision signals to the 3D eye parameter estimation. 
An overview of the described pipeline can be found in Figure~\ref{fig: Flow}. 

\subsection{Network Architecture}

The input eye image is denoted as $I \in \mathbb{R}^{H \times W \times C}$, where $H$ is the image height and $W$ is the width.
The image is represented by $C$ color channels. 
A backbone feature extractor $\mathsf{B}(\cdot)$ takes the image $I$ as the input and produces a feature map $F = \mathsf{B} (I)$.
Usually $F \in \mathbb{R}^{\frac{H}{s} \times \frac{W}{s} \times D}$ has $D$ channels, and the original spatial dimensions are downscaled by a factor of $s$. 
The feature map $F$ can be directly used for various computer vision tasks (e.g. classification, regression, dense predictions, etc.).
In our approach, the spatial dimensions are collapsed with global average pooling to obtain $F \in \mathbb{R}^{1 \times 1 \times D}$, representing the extracted global features of each eye image.
We choose ResNet-50~\cite{he2016residual} as our backbone feature extractor $\mathsf{B}(\cdot)$.

At this stage, the sequence of consecutive eye video frames $\mathcal{I} = \{I_1, ..., I_n, ..., I_N\}$ is transformed into a sequence of global eye features $\mathcal{F} = \{F_1, ..., F_n, ..., F_N\}$ independently. Next, we jointly process the sequence of eye features $\mathcal{F}$ to estimate 3D eye parameters $\mathcal{E} = \{E_1, ..., E_n, ..., E_N\}$ for each eye frame. We use a joint-processing network $\mathsf{T}(\cdot)$, which takes the ordered sequence of eye features $\mathcal{F}$ and outputs the corresponding eye parameters $\mathcal{E}=\mathsf{T}(\mathcal{F})$. 
Note that some eye parameters do not change across frames of the same individual (e.g. eyeball radius, iris radius), whereas some eye parameters are different for each frame (e.g. gaze direction, pupil radius). Nevertheless, the different eye parameters are very correlated for consecutive frames. Through joint processing,  valuable information is shared between consecutive frames.
Specifically, we choose the popular transformer encoder architecture~\cite{vaswani2017transformers} as our joint processing network $\mathsf{T}(\cdot)$. It has been shown that applying a transformer on independently extracted features of consecutive frames is very effective for per-frame video predictions~\cite{wang2020end}. Our proposed network architecture is visualized in Figure~\ref{fig: Flow}.

\begin{figure}[t!]
    \centering
    \begin{subfigure}{\columnwidth}
        \centering
        \includegraphics[width=0.95\columnwidth]{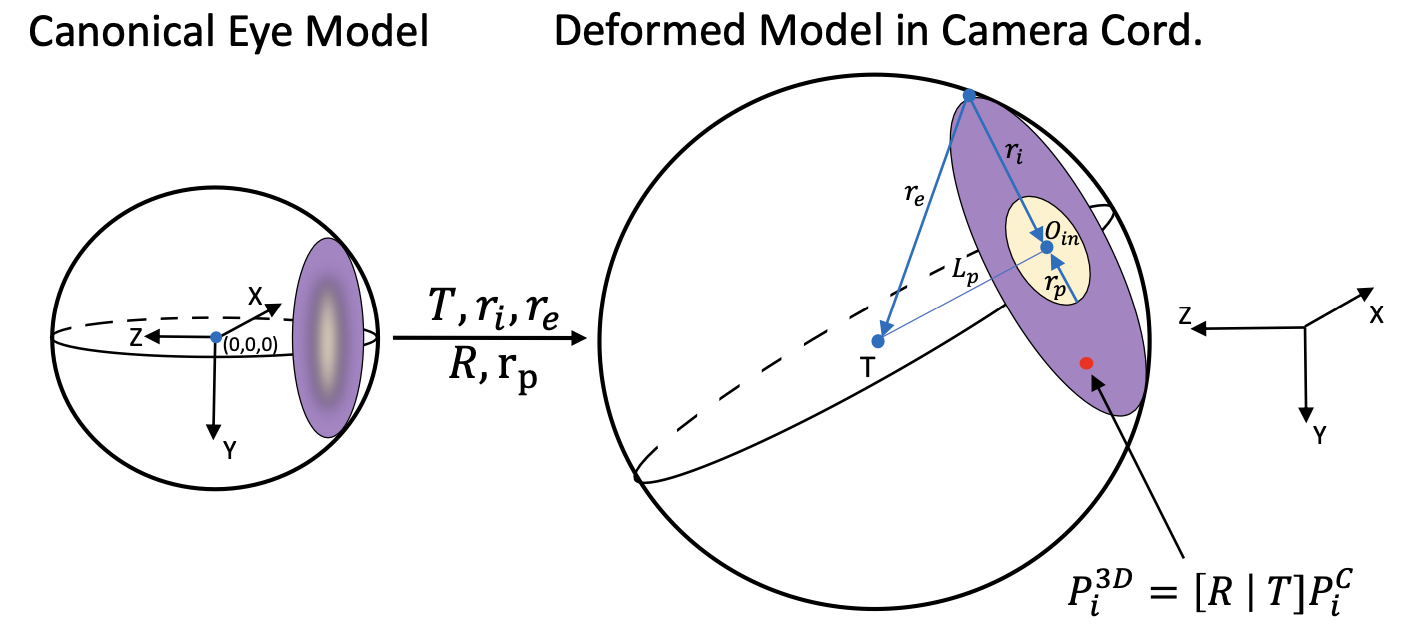}
        \caption{{Using the predicted eye parameters, we obtain the deformed eye model in the camera coordinate frame.} 
        }
        \label{fig:deformed_eye}
    \end{subfigure}
    \begin{subfigure}{\columnwidth}
        \centering
        \includegraphics[width=0.95\columnwidth]{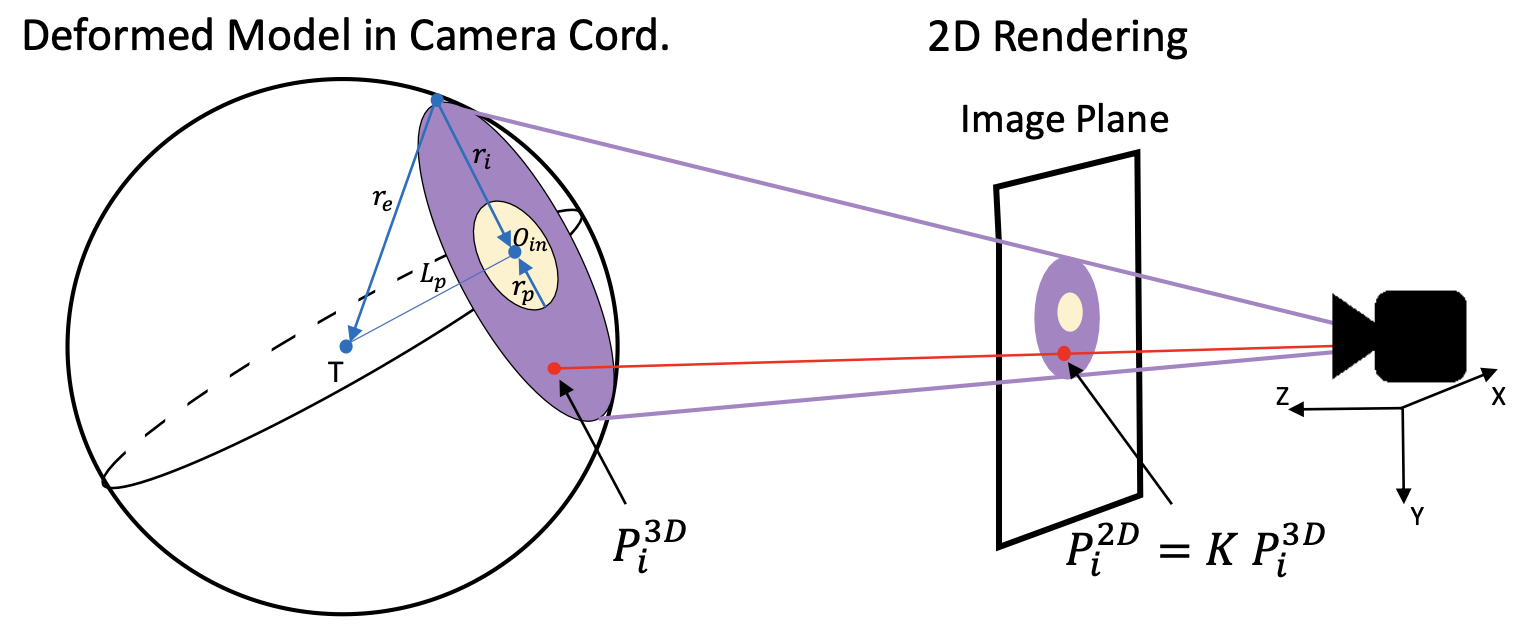}
        \caption{{We project the iris and pupil regions of the deformed eye to the image plane and obtain the corresponding 2D segmentation masks.}
        }
        \label{fig:2d_rendering}
    \end{subfigure}
    \vspace{-0.7cm}
    \caption{A sequence of differentiable operations to render 2D semantics as a function of the predicted camera and eye parameters. 
    }
    \vspace{-2pt}
    \label{fig:eye_model}
\end{figure}

\subsection{Deformable Eye Model}
Similar to previous work~\cite{dierkes2018novel, Swirski2013AF}, we model the eyeball as a sphere defined by its center $o_e$ and radius $r_e$. A second sphere intersects with the eyeball and the points of intersection form the iris circle defined by its center $o_i$ and radius $r_i$. Inside the iris circle, there is a concentric pupil circle with $o_p=o_i$ and radius $r_p$. The normalized optical axis $g$ can be obtained as the vector through the two centers $g = \frac{o_i - o_e}{ \| o_i - o_e \|}$. We consider $g$ the approximated gaze vector.
Note that we do not model the $\kappa$ angle offset between the optical and the visual axes.
The described eye model can be observed in Figure~\ref{fig:deformed_eye}. 

For each frame $I_n$, we estimate the following eye parameters 
$E_n=\{r_{e}, r_{i}, r_{pn}, T, R_{n}\}$. 
Eyeball radius $r_e$, iris radius $r_i$, and the eyeball translation $T$ are estimated to be the same for all consecutive frames in the sequence. 
$T$ is the translation vector from the eyeball center to the camera center.
In other words, $T$ is the eyeball center position expressed in camera coordinates. 
Furthermore, eyeball rotation $R_n$ and pupil radius $r_{pn}$ are estimated differently for each frame. 
$R_n$ is a rotation matrix that rotates the camera's coordinate system such that the negative z-axis becomes co-linear with the gaze vector $g_n$ and shares the same direction. The camera coordinate system has a positive x-axis pointing to the right, a positive y-axis pointing down, and positive z-axis pointing in front of the camera. Therefore, the negative z-axis naturally looks towards the camera and the optical axis can be computed as $g_n=R_n[0,0,-1]^T$ in camera coordinates. 
Due to human eye movement constraints,  
the eyeball rotation matrix is constrained to only allow for pitch and yaw of maximum $80 ^{\circ}$ (the roll is always zero).
All other parameters of the eyeball model can be computed, if needed, from the estimated five parameters $\{r_{e}, r_{i}, r_{pn}, T, R_{n}\}$. For example, the pupil and iris center can be calculated as $o_{in} = o_{e} + L_{p} g_n$, where $L_{p} = \sqrt{r_{e}^2 - r_{i}^2}$ is the distance from the eyeball center to the pupil center.

In addition, we also estimate the camera's intrinsic parameters, which are usually not provided in publicly available datasets. 
They are estimated to be the same for the whole sequence since all frames were captured with the same camera. More specifically, we use the pinhole camera model with the same focal length $f_x = f_y = f$ and assume the camera center to be $(c_x, c_y)=(\frac{W}{2}, \frac{H}{2})$.

In order to weakly supervise the 3D eye parameters with the 2D semantic labels in a fully differentiable manner, we generate discrete point clouds for the pupil and iris of the 3D eye model. We first create a canonical eye model which is defined by $o_e^{C}=(0,0,0)$ and $g^{C}=(0,0,-1)$. The eyeball is positioned in the center of the canonical coordinate system and the optical axis is co-linear with the canonical z-axis and points to the negative direction. 
Also, the canonical coordinate system shares the same rotation as the camera's coordinate system.
Next, we generate a discrete point cloud for the pupil circle and iris disk, based on the estimated eye parameters,

\begin{equation}
P_p^{C}=\{ \left[r_p \rho cos(\theta), r_p \rho sin(\theta), -L_p\right] | \rho \in [0,1], \theta \in [0, 2\pi) \},
\end{equation} 

\begin{equation}
\begin{aligned}
P_i^{C}=\{ \left[r cos(\theta), r sin(\theta), -L_p\right] | &\\ r=r_p + \rho (r_i - r_p), &\rho \in [0,1], \theta \in [0, 2\pi) \}.
\end{aligned}
\end{equation} 
We then obtain the deformed point clouds in the camera coordinate system as follows:
\begin{equation}
    P_p^{3D} = [R | T] P_p^{C}.
\end{equation}
\begin{equation}
    P_i^{3D} = [R | T] P_i^{C}.
\end{equation}
This deformation process is depicted in Figure~\ref{fig:deformed_eye}.
Finally, we project the deformed 3D iris and pupil point clouds onto the 2D camera screen,
\begin{equation}
    P_p^{2D} = K P_p^{3D} = \left[ \begin{array}{ccc} f & 0 & W/2\\ 0 & f & H/2 \\ 0 & 0 & 1\end{array} \right] P_p^{3D}.
\end{equation}
The same equation is applied to obtain $P_i^{2D}$. This rendering process is depicted in Figure~\ref{fig:2d_rendering}.

\subsection{Loss functions}
In order to supervise the produced point clouds with semantic labels, we design a two-part loss which minimizes the Euclidean distance between the projected semantic point clouds $P^{2D}$ and the ground truth semantic masks $P^{GT}$. 
Firstly, for each frame $n$ and point $k$ in the point cloud $P_{nk}^{2D}$, the Euclidean distance is computed with respect to its closest pixel position in the ground truth masks,
\begin{equation}
    L_{pred2GT} = \frac{1}{NK} \sum_{n=1}^{N} \sum_{k=1}^{K} \|  P_{nk}^{2D} - p_{n \arg\min_{j} \|P_{nk}^{2D} - p_{nj}^{GT}\|}^{GT} \|.
\end{equation}
Also, to avoid unpenalized pupil/iris shrinking, we introduce the same loss in the opposite direction,
\begin{equation}
    L_{GT2pred} = \frac{1}{NK} \sum_{n=1}^{N} \sum_{k=1}^{K} \|  P_{nk}^{GT} - p_{n \arg\min_{j} \|P_{nk}^{GT} - p_{nj}^{2D}\|}^{2D} \|.
\end{equation}
The defined losses are applied to both the pupil and iris point clouds with weights $w_{pred2GT}$ and $w_{GT2pred}$,
\begin{equation}
    \begin{aligned}
    L_{seg} = w_{pred2GT} (L_{pred2GT}^{p} + L_{pred2GT}^{i}) \\ + w_{GT2pred} (L_{GT2pred}^{p} + L_{GT2pred}^{i}).
    \end{aligned}
\end{equation}
The objective of these losses is to guide the learning process to generate better 3D eye model parameters such that the projected iris and pupil match the observed 2D semantic masks.

When supervising the estimated gaze vector, we use the mean square error loss with weight $w_{gaze}$

\begin{equation}
    L_{gaze} =  w_{gaze}  \frac{1}{N}\sum_{n=1}^{N} \| g_n - g_n^{GT}\| .
\end{equation}

Also, in some additional experiments we supervise the projected eyeball center $o_e^{2D} = K o_e^{3D}$ with weighted Euclidean distance loss,

\begin{equation}
    L_{center} =  w_{center}  \frac{1}{N}\sum_{n=1}^{N} \| o_e^{2D} - o_e^{GT}\| .
\end{equation}


%% file: latex/4_experiments.tex
\section{Experiments}

\subsection{Experimental setup}
\myparagraph{Dataset}
The TEyeD~\cite{fuhl2021teyed} dataset contains more than 20 million eye images captured by head-mounted devices with near-eye infrared cameras. The dataset was made from 132 participants, who recorded multiple different videos while they performed various indoor and outdoor tasks which cover a wide range of realistic eye movements. 
This dataset provides ground truth labels for 3D gaze vectors, 2D segmentation masks, landmarks for both pupil and iris, as well as other event annotations such as blink. 
TEyeD does not have a predefined data split, so we randomly select around 348K images for training and around 36k images for testing. 
We generate the fixed train/test splits by randomly selecting frames from different recordings, such that frames from one recording can exclusively be either in the training or the test split.
Additionally, temporal downsampling is applied to reduce the frame rate from 25 Hz to 6.25 Hz, so that there is significant eye movements and to avoid identical eye images.  
To the best of our knowledge, no other dataset with both 3D gaze labels and semantic segmentation masks exists.  

\myparagraph{Baseline}
The standard appearance-based approaches directly predict gaze vectors from a neural network. Similarly, we consider the joint-processing network $\mathsf{T}(\cdot)$ as a baseline, which directly estimates gaze vectors without modeling the 3D eyes.  

\myparagraph{Implementation details}
Our experiments focus on glint-free gaze estimation from infrared near-eye video frames. 
The joint-processing network $\mathsf{T}$ has 3 encoder blocks with an embedding dimension of 256, 8 attention heads, an MLP expansion ratio of 2, and no dropout.
Moreover, we optimize our model with the LAMB optimizer~\cite{lamb} along with a cosine learning rate scheduler with a warm-up of 16K iteration. 
The initial learning rate is 2e-3, and the cosine scheduler gradually drops to 2e-5 over 320k training iterations.
The experiments use a batch size of 4, where each batch contains 4 consecutive eye video frames. 
When generating iris and pupil point clouds, we uniformly sample 72 angles in the range $\theta \in [0, 2\pi)$ for 8 radius values in the range $\rho \in [0, 1]$, to form a circle template for the pupil and a disc template for the iris with each having 576 points. 
Furthermore, to mitigate overfitting, the training dataset is additionally augmented by applying 1.0-2.0 standard deviation blur,  0-30\% random noise, and a horizontal flip of the eye image with 20\% probability. 
Moreover, for the few-shot learning experiments, we use a much smaller portion of the training data. We also reduce the number of iterations to 80K, which is shown to be more than sufficient for convergence.
Starting from the pre-trained weights, the initial learning rate is lowered to 2e-4 and gradually reduced down to 2e-6 with the cosine scheduler. The number of iterations is reduced to 30K.  
All few-shot experiments are evaluated on the complete test set (36K).

\subsection{Results}

\begin{table}[t!]
\centering
\small
\vspace{-4pt}
\caption{When training with semantic loss only, the estimated model renders good semantic regions, but not good gaze, due to the ill-posed problem. Adding more 3D constraints improves the 3D eye model estimation. The gaze-only baseline estimates very good gaze vectors but does not offer anything in addition.
} 
\begin{tabular}{|c|c|c|c|c|}
\hline
 \makecell{Loss} & 
 \makecell{3D gaze \\ {[\textdegree]} $\downarrow$} & 
 \makecell{2D gaze \\ {[\textdegree]} $\downarrow$} &
 \makecell{Sem. \\ mIoU $\uparrow$} & 
 \makecell{2D eye \\ cent. {[px]} $\downarrow$}  \\
 \hline \hline
 
\makecell{Gaze \\ (concat.)~\cite{fuhl2021teyed}}
& 7.10 & 61.74 & N/A & N/A \\  \hline 
\makecell{Gaze \\ (ours)}
& 0.96 & 6.93 & N/A & N/A \\  \hline \hline
Sem.
& 20.12 & 39.09 & 93.0 \% & 11.41 \\  \hline

Sem. + Gaze
& 0.99 & 7.42 & 92.8 \% & 6.65 \\  \hline \hline
\makecell{Sem. + Gaze \\ + Cent.}
& 1.21 & 10.39 & 92.4 \% & 2.02 \\  \hline 

\end{tabular}
\label{table:general_behaviour}
\end{table}

\begin{figure}[t!]
    \centering
    \begin{subfigure}{\columnwidth}
        \centering
        \includegraphics[width=0.9\columnwidth]{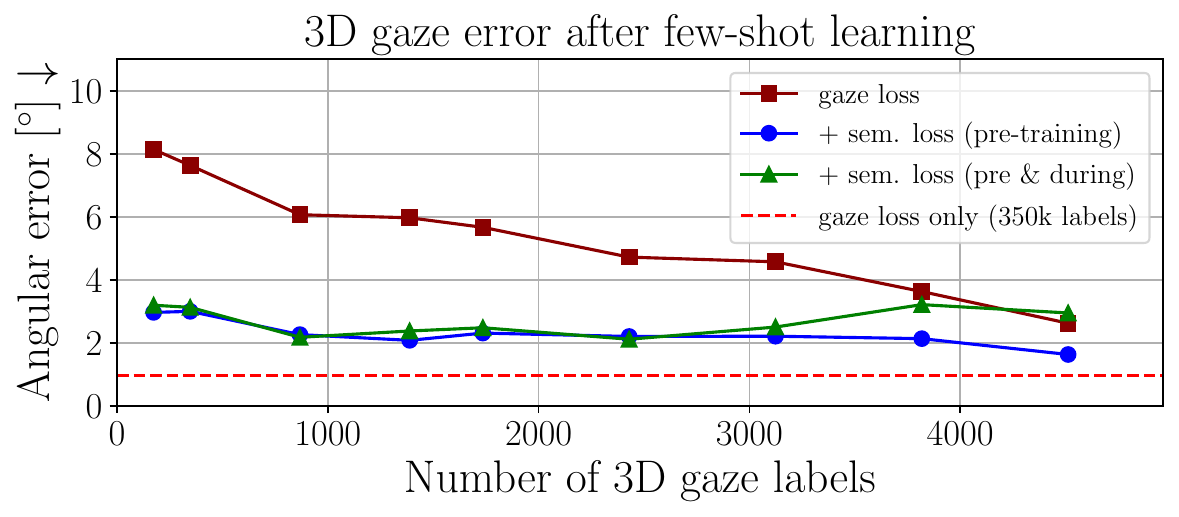}
        \label{fig:few_shot_3D}
    \end{subfigure}
    \vspace{-0.5cm}
    \begin{subfigure}{\columnwidth}
        \centering
        \includegraphics[width=0.9\columnwidth]{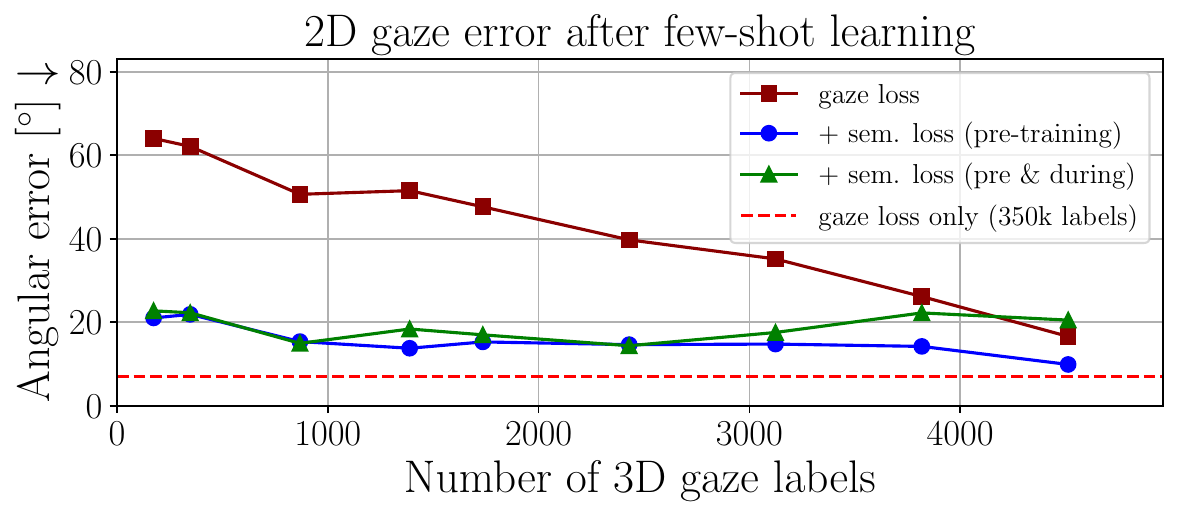}
        \label{fig:few_shot_2D}
    \end{subfigure}
    \vspace{-0.15cm}
    \caption{Few shot learning experiments, where models are trained on small amounts of 3D gaze labels. Supervising from scratch is difficult for the model. However, fine-tuning from a model supervised with large amounts of semantic labels facilitates gaze prediction.}
    \vspace{-5pt}
    \label{fig:few_shot}
\end{figure}

\begin{figure}[t!]
    \centering
    \includegraphics[width=0.9\columnwidth]{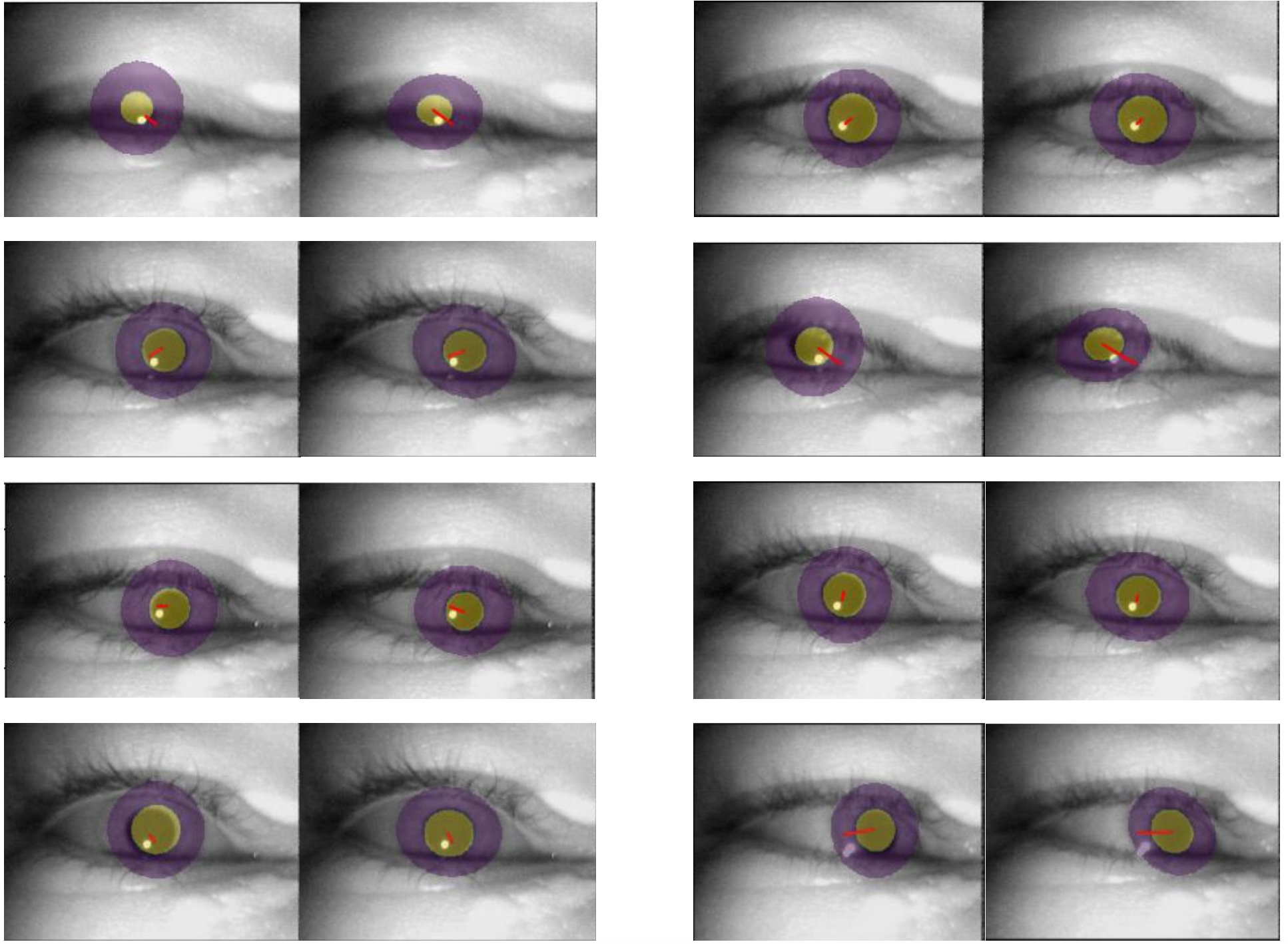}
    \caption{Qualitative results of a few shot supervision (0.05\% of available labels), after pre-training the network with a huge amount of semantic labels. The ground eyeball center was used while visualizing the projected estimated gaze vector.}
    \label{fig:few_shot_qualitative}
    \vspace{-0.5cm}
\end{figure}

\begin{figure}
    \centering
    \includegraphics[width=0.9\columnwidth]{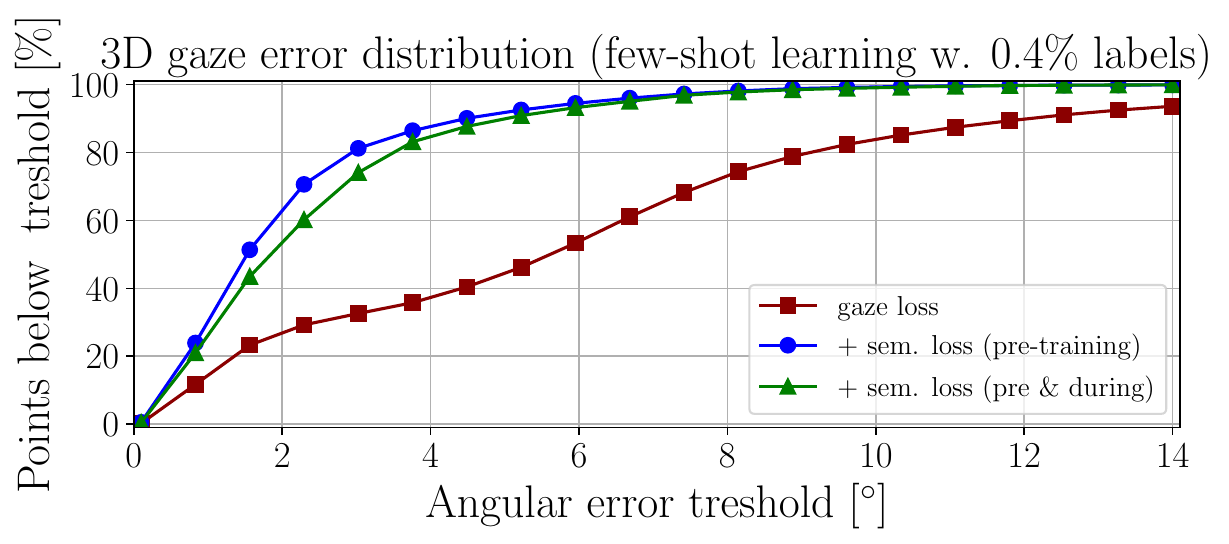}
    \vspace{-5pt}
    \caption{The cumulative distribution of error (in angles) when predicted 3D eye gaze vectors after few-shot fine-tuning, for different methods reported in Figure~\ref{fig:few_shot}.}
    \label{fig:cumulative_error_few_shot}
    \vspace{-0.5cm}
\end{figure}
Firstly, we test our proposed network architecture by supervising only using gaze labels. The supervision is performed using the whole training split. We compare our network to the gaze estimation solution proposed in~\cite{fuhl2021teyed}. This solution immediately concatenates all the sequence frames before passing them into the ResNet-50 backbone, therefore there is no joint processing step. In Table~\ref{table:general_behaviour} we clearly see that our architecture outperforms the model used in~\cite{fuhl2021teyed}.

Next, we examine the general behavior of supervising our method on the complete training split. The results are presented in Table~\ref{table:general_behaviour}. When only the semantic loss is used, our method reconstructs 3D eye models which render very accurate semantic regions of the pupil and iris. However, the quality of the estimated gaze vectors and the projected eyeball centers is not good. This is because the problem is ill-posed since there exist multiple configurations of the eye in 3D which render the same 2D semantics. For example, imagine a fixed pupil point cloud that renders a perfect 2D pupil mask. There exist many corresponding valid eyeballs, with different centers and radii. All of those different models produce different gaze vectors, but the same pupil mask. To complicate things more, more plausible configurations exist with unknown camera intrinsics.
Therefore, a good learning strategy must impose more constraints and supervision on the 3D model. 
Now, when the gaze is supervised along with the semantics, the results are much better. The model still renders very accurate semantic regions, however, it also produces good gaze and eyeball center estimates. This is because the additional gaze label provides more constraints and supervision to the 3D eye model. 
If we additionally supervise the projected eyeball center, we achieve similar semantic and gaze performance, with a slight increase in eyeball center estimation performance. 
Finally, when directly predicting the gaze vector only, without an eye model (like many appearance-based approaches), the network achieves very good gaze results. However, apart from the estimated gaze, this network does not offer anything in addition, unlike our method which provides the complete 3D eye model.

Utilizing both semantic and gaze loss simultaneously would be an ideal approach; however, there is a significant challenge associated with this method. 
3D gaze labels for head-mounted devices are very rare and scarce in publicly available datasets~\cite{fuhl2021teyed} and they are also very difficult to annotate for newly collected data. On the other hand, there is an abundance of semantic iris and pupil labels in publicly available datasets~\cite{fuhl2021teyed} and they are very easy to annotate for newly collected data. Therefore, the widely available semantic labels can be utilized to train a model from scratch, to serve as a good starting point. The network weights of such a model can then be fine-tuned on a small amount of available gaze labels, in order to impose more 3D supervision and constraints.
Figure~\ref{fig:few_shot} contains few-shot learning experiments, where models are trained only with a small amount of 3D gaze labels.
Supervising from scratch with only the gaze loss with a small number of labels is difficult for the network. However, fine-tuning a network that was previously supervised with many semantic labels achieves much better performance and facilitates gaze vector estimation.
Figure~\ref{fig:few_shot_qualitative} depicts qualitative results of a few-shot fine-tuning training. Also, the error distribution of estimating 3D gaze with few-shot fine-tuning is shown in Figure~\ref{fig:cumulative_error_few_shot}.



%% file: latex/5_conclusion.tex
\section{Conclusion}
We propose a hybrid method for 3D gaze estimation by considering a deformable 3D eye model, taking advantage of both appearance-based and model-based approaches.  
Our method also predicts corresponding camera parameters, allowing us to project the 3D eye model onto the image plane.
Specifically, we propose a differentiable eye model that selects a dense set of 3D points with known semantics from the canonical eye model, followed by deformation and projection onto the image plane. 
The projected points are then compared against the provided 2D segmentation masks, which serve as weak labels during the whole process. 
In addition, we also make use of the supervision of 3D gaze. 
Our experimental evaluations clearly demonstrate the benefits of the proposed method in learning 3D eye gaze, from video frames, using the joint supervision considered on the practical grounds. 
Thanks to the model-aware weak supervision of the segmentation masks, fewer 3D gaze labels are needed. 
The recovered eye model may possibly be used beyond the 3D gaze estimation task. 
More importantly, our differentiable eye model may be used beyond the context of this paper.
